\documentclass[]{spie}  %>>> use for US letter paper
\usepackage{amsmath,amsfonts,amssymb}
\usepackage{graphicx}
\usepackage[colorlinks=true, allcolors=blue]{hyperref}
\usepackage{colortbl}
\usepackage{multirow}
\usepackage{booktabs}
\usepackage{subcaption}
\title{An evaluation of large pre-trained models \\ for gesture recognition using synthetic videos}

\author[a,b]{Arun Reddy}
\author[b]{Ketul Shah}
\author[a]{Corban Rivera}
\author[a]{William Paul}
\author[c]{Celso M. De Melo}
\author[b]{Rama Chellappa}

\affil[a]{Johns Hopkins University Applied Physics Laboratory, Laurel, MD, USA}
\affil[b]{Johns Hopkins University, Baltimore, MD, USA}
\affil[c]{Army Research Laboratory, Los Angeles, CA, USA}
\begin{document} 
\maketitle

\begin{abstract}
 In this work, we explore the possibility of using synthetically generated data for video-based gesture recognition with large pre-trained models. We consider whether these models have sufficiently robust and expressive representation spaces to enable ``training-free" classification. Specifically, we utilize various state-of-the-art video encoders to extract features for use in k-nearest neighbors classification, where the training data points are derived from synthetic videos only. We compare these results with another training-free approach--- zero-shot classification using text descriptions of each gesture. 
 In our experiments with the RoCoG-v2 dataset, we find that using synthetic training videos yields significantly lower classification accuracy on real test videos compared to using a relatively small number of real training videos. We also observe that video backbones that were fine-tuned on classification tasks serve as superior feature extractors, and that the choice of fine-tuning data has a substantial impact on k-nearest neighbors performance.
Lastly, we find that zero-shot text-based classification performs poorly on the gesture recognition task, as gestures are not easily described through natural language.
 % We observe that vision-language self-supervised models perform better than unimodal self-supervised models. Fine-tuning models leads to improvements in performance, however, it is important to choose the fine-tuning dataset based on the characteristics of the target set for best results.     
\end{abstract}


\section{INTRODUCTION}
\label{sec:intro}  % \label{} allows reference to this section

One of the promises of synthetic data is the possibility of reducing reliance on real data for training deep learning models, which can introduce practical challenges and ethical concerns. As such, there has been a growing interest in using synthetically generated video data to train models for various video-related tasks. However, previous work has shown the existence of a large domain gap between real and synthetic video data, resulting in sub-optimal performance when naively applying a synthetically-trained model to data from the real domain \cite{reddySynthetictoRealDomainAdaptation2023, shah2023multi, da2022dual, katyalLeveragingSyntheticData2023}. This issue has spurred the development of various approaches for video domain adaptation 
\cite{reddyUnsupervisedVideoDomain2024, da2022dual, wei2023transvae}, which can be computationally expensive and difficult to implement in practice.

Here, we consider whether state-of-the-art video backbones, given the scale of their pre-training, are capable of extracting domain-invariant representations to enable video classification without needing any real data for the task. Specifically, we experiment with two different ``training-free" approaches. The first uses modern video backbones as feature extractors for K-nearest neighbors (KNN) classification, where the training samples are derived from synthetic videos. In the second, we rely on text descriptions of the classes and attempt to perform zero-shot classification using similarity between video features and text features from each class description. We use video-based gesture recognition as our video classification task in this study.

\begin{figure}[h!]
    \centering
    \includegraphics[scale=0.18]{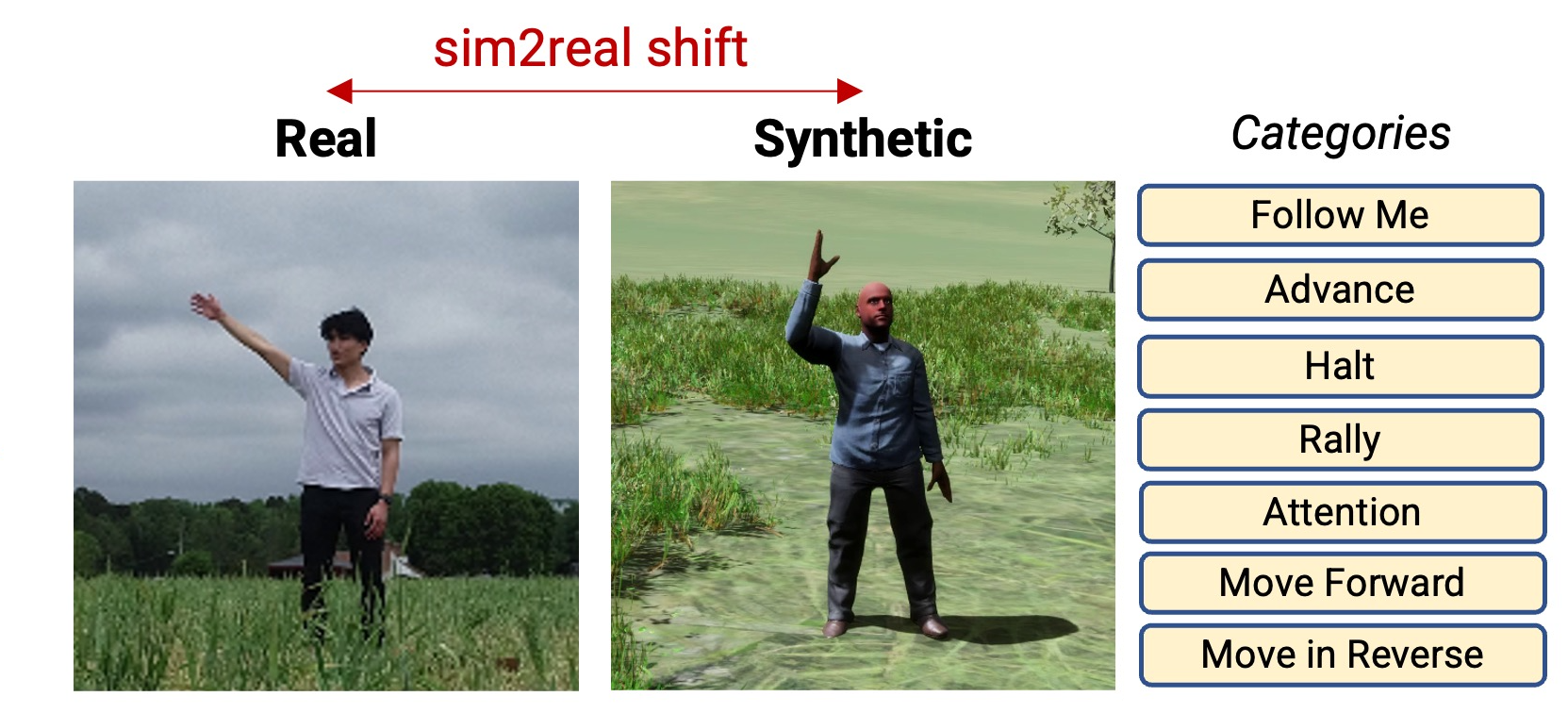}
    \caption{Examples of a real (left) and synthetic (right) video from RoCoG-v2. The datset consists of 7 gesture categories.}
    \label{fig:rocog-examples}
\end{figure}

\section{EXPERIMENTS}
\label{sec:exp}  

% In this work, we study whether a gesture recognition task can be specified without collecting real-world data. In particular, we consider two ways to provide information about the gesture classes: (1) using synthetic videos, and (2) using textual descriptions of gestures. 
We perform two sets of experiments. In the first set, we leverage large pre-trained video models as feature extractors and perform KNN classification using these features. In the second, we perform zero-shot classification using textual descriptions of gestures. These two methods are described in detail below. 
All experiments are performed on ground viewpoint videos from the RoCoG-v2 dataset \cite{reddySynthetictoRealDomainAdaptation2023}, examples of which are shown in Figure 1.  

\subsection{KNN Classification}
When using synthetic training videos to specify the gesture recognition task, we perform KNN classification on features extracted from large pre-trained models. More specifically, the synthetic training data (44K videos) is embedded in the feature space of a video encoder, and for a given real test video, the majority vote of the K nearest training data points is used for classification, based on L2 distance. We also experiment with a scenario where we have a small real dataset (203 videos) available for training. We set K=3 for all experiments. We consider three different types of video encoders based on pre-training (and fine-tuning) strategies, as described below. We choose a ViT-B/16 model for all experiments, and also study the effect of a larger ViT-L/16 model for the best setting.    

\noindent\textbf{Self-Supervised Pre-Training.} We consider Unmasked Teacher (UMT) \cite{liUnmaskedTeacherTrainingEfficient2023}, which is a state-of-the-art video self-supervised learning approach. This approach masks out most of the video tokens and enforces alignment between the representations of unmasked patches and the corresponding ones from a teacher model (CLIP \cite{radford2021learning}). We use the UMT model pre-trained on K710 videos (a union of K400 \cite{kay2017kinetics}, K600 \cite{carreiraShortNoteKinetics6002018} and K700 \cite{carreira2019short}) for our experiments. Eight frames are sampled from each video using the TSN \cite{wang2018temporal} frame-sampling strategy. The entire video is divided into eight segments, and one frame is selected at random from each segment. The input frames to the network are resized to 224 $\times$ 224 resolution.       

\noindent\textbf{Vision-Language Pre-Training.} In contrast to UMT which uses only video data for pre-training, here we consider a vision-language pre-training approach of ViCLIP \cite{wangInternVidLargescaleVideoText2024}. ViCLIP uses a video-language contrastive objective similar to CLIP \cite{radford2021learning}, while also masking videos for efficient pre-training. We use the model pre-trained on a filtered version of the InternVid~\cite{wangInternVidLargescaleVideoText2024} dataset that has 10M video-text pairs. Eight frames of 224 $\times$ 224 resolution are used as input to the network.        

\noindent\textbf{Self-Supervised Pre-Training + Supervised Fine-Tuning.} Here, we use self-supervised models which were further fine-tuned for video classification in a supervised manner. We consider two pre-training methods, UMT \cite{liUnmaskedTeacherTrainingEfficient2023} and VideoMAE~\cite{tongVideoMAEMaskedAutoencoders2022}. VideoMAE is a powerful self-supervised pre-training approach which works by encoding partially masked inputs and reconstructing the masked out regions. The VideoMAE models are pre-trained on a larger (1.35M) UnlabeledHybrid~\cite{wang2023videomae} dataset, whereas UMT models are pre-trained on the K710 dataset (650K). These models are either fine-tuned on Kinetics~\cite{kay2017kinetics} (K710, K400, K600, K700) or the more motion-centric Something-Something-v2 (SSv2)~\cite{goyal2017something} dataset. For the VideoMAE models, we sample sixteen frames from the input video, with a tubelet size of two frames, resulting in the same number of tokens as using eight frames with a tubelet size of one frame, as in the above scenarios.

\begin{table*}[ht]
\small
\centering
\resizebox{1.0\linewidth}{!}{
\begin{tabular}{c|c|cc|c|cc}
\toprule
\multirow{2}{*}{\textbf{Type}} & \multirow{2}{*}{\textbf{Backbone}} & \textbf{Pre-Training} & \textbf{Pre-Training} & {\textbf{Fine-Tuning}} & \multicolumn{2}{c}{\textbf{Real Test KNN Acc. (\%)}} \\
& & \textbf{Method} & \textbf{Data} & {\textbf{Data}} & Synthetic Train & Real Train  \\
\midrule

Self-Supervised Pre-Training & ViT-B/16 &      UMT & K710              & -           & 18.2 & 31.2 \\
\midrule
Vision-Language Pre-Training & ViT-B/16 &   ViCLIP & InternVid FLT-10M & -           & 19.2 & 40.4 \\
\midrule
                             & ViT-B/16 &      UMT & K710              & K710        & 42.4 & 49.5 \\
                             & ViT-B/16 &      UMT & K710              & K710 + K400 & 38.4 & 45.5 \\
Self-Supervised Pre-Training & ViT-B/16 &      UMT & K710              & K710 + K600 & 33.3 & 49.5 \\
+ Supervised Fine-Tuning     & ViT-B/16 &      UMT & K710              & K710 + K700 & 35.4 & 51.5 \\
                             & ViT-B/16 & VideoMAE & UnlabeledHybrid   & K710        & 32.3 & 60.6 \\
                             & ViT-B/16 & VideoMAE & UnlabeledHybrid   & SSv2        & 43.4 & 68.7 \\
                             & ViT-L/16 & VideoMAE & UnlabeledHybrid   & SSv2        & 64.6 & 71.7 \\

\bottomrule
\end{tabular}}
\vspace{5pt}
\caption{K-nearest neighbor classification results on RoCoG-v2 ground videos using a variety of video backbones.}

\label{tab:knn_results}
\end{table*}

\subsection{Zero-Shot Text-Based Classification}
% given a text description and encoders, how do we perform classification  
% two different kinds of text descriptions 
The gesture recognition task can alternatively be specified by providing the textual descriptions of each activity. In our second set of experiments, shown in Table 1, we use text descriptions to perform zero-shot classification using pre-trained vision-language models. Specifically, we use the pre-trained ViCLIP \cite{wangInternVidLargescaleVideoText2024} text and video encoders, where classification is performed based on similarity between video features and text embeddings of descriptions of all classes. We use two kinds of text descriptions of the gestures, original and transformed, as follows:  

\noindent\textbf{Original.} These are the instructions associated with each gesture as they appear in the US Army Field Manual \cite{USArmy87}. These were provided to the performers of the gestures in the RoCoG-v2 dataset.

\noindent\textbf{Transformed.} We use GPT-3.5 to turn the original text instructions into gesture descriptions by prompting the model with the following text: ``\textit{Can you summarize the description below of an activity instruction and start with ``A person"}". 

\begin{table*}[ht]
\small
\centering
\resizebox{0.6\linewidth}{!}{
\begin{tabular}{c|c|c|c}
\toprule
\multirow{1}{*}{\textbf{Model}} & \multirow{1}{*}{\textbf{Training Data}} & \textbf{Text Description} & {\textbf{Test Accuracy (\%)}} \\

\midrule

ViCLIP-B & InternVid FLT-10M & Original & 25.3 \\

ViCLIP-B & InternVid FLT-10M & Transformed & 26.3 \\

\bottomrule
\end{tabular}}
\vspace{5pt}
\caption{Zero-shot classification on RoCoG-v2 ground videos using two forms of text descriptions.}

\label{tab:zs_results}
\end{table*}

\section{DISCUSSION}
\label{sec:discussion}  % \label{} allows reference to this section

Several observations can be made from the KNN classification results in Table 1. First, we can clearly see that, in all cases, using synthetic training data results in lower accuracy on real videos than using real training data. Despite the large quantity of synthetic training data (roughly 200$\times$ that of real training data), we find it is not as effective at defining the gesture classes in the RoCoG-v2 dataset as real data. This is indicative of the synthetic-to-real domain gap that has been observed previously, which is still far from solved. Figure 2 also illustrates this domain gap, as real video features appear noticeably more clustered by gesture class than synthetic video features.

\begin{figure*}[t!]
    \centering
    \begin{subfigure}[t]{0.4\textwidth}
        \centering
        \includegraphics[scale=0.4]{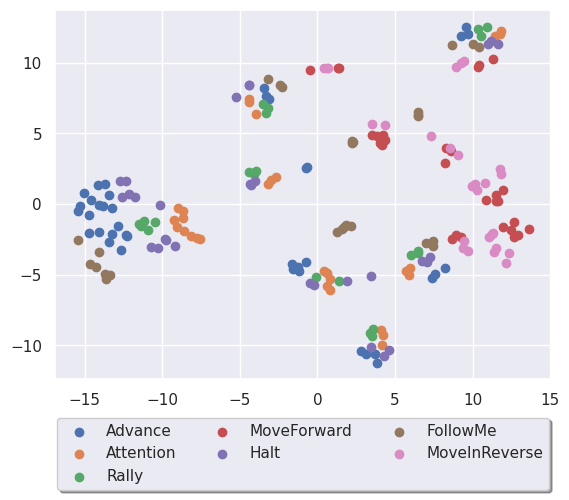}
        \caption{t-SNE plot for real data features.}
    \end{subfigure}%
    ~ 
    \begin{subfigure}[t]{0.4\textwidth}
        \centering
        \includegraphics[scale=0.4]{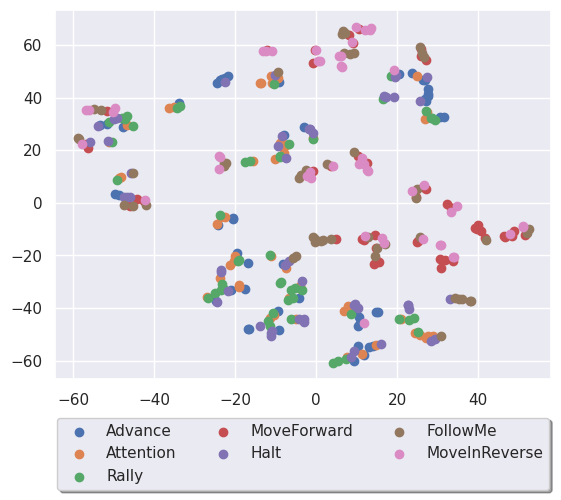}
        \caption{t-SNE plot for synthetic data features.}
    \end{subfigure}
    \vspace{5pt}
    \caption{t-SNE plots for real and synthetic data. Real data is more meaningfully clustered compared to synthetic data, as the features are extracted using a ViT-B/16 model pre-trained on K710 and fine-tuned on K710 and K400. For (a), we use all real data whereas for (b), we use 50 samples per class chosen at random from the synthetic dataset.}
\end{figure*}
\label{fig:tsne}

Next, we can see from comparing the first two rows in Table 1 that large-scale vision-language pre-training seems to confer some benefit over smaller-scale masked pre-training, particularly when using real videos for KNN classification. However, we find that backbones that have undergone supervised fine-tuning vastly outperform both the self-supervised and vision-language pre-trained backbones. Interestingly, as the backbone is fine-tuned on more real videos, we see a drop in KNN accuracy using synthetic training data while accuracy using real training data increases. This suggests that video transformer backbones become less robust to the synthetic-to-real domain shift the more they are trained on real videos. 

We find that the choice of fine-tuning data has a substantial impact on KNN classification. We can see that backbones fine-tuned on the SSv2 dataset \cite{goyal2017something} perform much better than those trained on Kinetics videos. SSv2 is a temporally-heavy dataset, where modeling of motion is critical for solving the classification task. In contrast, the action categories in Kinetics videos exhibit a high degree of object and scene bias. Because the gesture recognition task in RoCoG-v2 is also motion-focused, SSv2 serves as an effective source of fine-tuning data. Notably, scaling up from the ViT-B model to ViT-L results in significant boosts in KNN accuracy, particularly when using synthetic training data. Future experiments should investigate whether, in general, larger models might be more robust to synthetic-to-real shifts.

Finally, we observe in Table 2 that zero-shot classification using ViCLIP \cite{wangInternVidLargescaleVideoText2024} performs poorly on the gesture recognition task, regardless of the type of text description used. This is likely because the gesture recognition task involves fine-grained motion differences that are not easily expressed through natural language. 

\acknowledgments % equivalent to \section*{ACKNOWLEDGMENTS}       
 
This research was sponsored by the Army Research Laboratory and was accomplished under Cooperative Agreement Number W911NF-21-2-0211. 
The views and conclusions contained in this document are those of the authors and should not be interpreted as representing the official policies, either expressed or implied, of the Army Research Office or the U.S. Government. 
The U.S. Government is authorized to reproduce and distribute reprints for Government purposes notwithstanding any copyright notation herein.

% References
\bibliography{report} % bibliography data in report.bib
\bibliographystyle{spiebib} % makes bibtex use spiebib.bst

\end{document}